\documentclass[conference]{IEEEtran}
\IEEEoverridecommandlockouts
\usepackage{cite}
\usepackage{amsmath,amssymb,amsfonts}
\usepackage{algorithmic}
\usepackage{graphicx}
\usepackage{textcomp}
\usepackage{xcolor}
\usepackage{lipsum}
\usepackage{multirow}
\def\BibTeX{{\rm B\kern-.05em{\sc i\kern-.025em b}\kern-.08em
    T\kern-.1667em\lower.7ex\hbox{E}\kern-.125emX}}

\begin{document}

\title{A Comprehensive Comparison of Deep Learning Architectures for COVID-19 Classification on CT \& X-ray Imagery\\}

\author{
\IEEEauthorblockN{Sarmad Khan}
\IEEEauthorblockA{\textit{Dept. of Computer \& Software Engg.}\\
\textit{Nat. Univ. of Sciences \& Technology}\\
Islamabad, Pakistan\\
rdxesss@gmail.com}
\and
\IEEEauthorblockN{Basim Azam}
\IEEEauthorblockA{\textit{School of Computing \& Info. Systems}\\
\textit{University of Melbourne}\\
Melbourne, Australia\\
basim.azam@unimelb.edu.au}
\and
\IEEEauthorblockN{Arslan Shaukat}
\IEEEauthorblockA{\textit{Dept. of Computer \& Software Engg.}\\
\textit{Nat. Univ. of Sciences \& Technology}\\
Islamabad, Pakistan\\
arslanshaukat@ceme.nust.edu.pk}
}

\maketitle

\begin{abstract}
COVID-19 was a significant challenge that led to the loss of numerous lives daily. Not only a certain country was involved in this outbreak, but even the world has suffered because of the coronavirus. Imaging techniques using computed tomography (CT) and X-rays of the lungs are the most useful tools for the COVID-19 or any other pandemic disease screening process. Technology today has revolutionized the world by using artificial intelligence to replace manual processes with automated machines, which enable the system to imitate the human brain by making wise decisions based on experience. Motivated by this, our work proposes to use convolutional neural networks (CNN) based models for designing a computer-aided diagnosis (CAD) system that differentiates between COVID-19 and healthy lung pictures. We used two different sets of X-ray images of the lungs in addition to two different sets of CT scans and the classification is done using a variety of networks that have been pre-trained such as VGG (16, 19), Densenet (121), Resnet (50, 50 V2, 101 V2), Mobile net (V2), Xception Inception (V3, Resnet V2), Efficient net (B0) and Nasnet (Large). On the X-ray and CT image datasets, Resnet and VGG architecture have shown the ability to properly differentiate COVID-19 from normal images, with an average accuracy of 95 to 98 percent respectively. Our acquired results on the classification datasets are competitive and superior to previously reported findings in the literature.
\end{abstract}


\section{Introduction}
The pandemic COVID-19, which began in late 2019, was without a doubt the most important event of 2020. Although some manufacturers have declared a vaccine, new strains have emerged, and the vaccine's impact on humans is unknown, necessitating further research. However, modern deep learning (DL) technologies can substantially aid physicians in diagnosing whether a person is infected with the coronavirus. \cite{chen2020deep} \cite{dias2020deeplms}.
\par
In recent years, several DNNs have been created as effective methods for addressing image-related challenges. Despite this, a major drawback of these artificial DNNs is the large amount of training data needed. These artificial DNNs should evaluate information and data to comprehend, as opposed to human individuals who can understand intelligently. As a result, when the model learns a few explicit frameworks, due to a lack of relevant data, overfitting may occur. Real-world scenarios often include imbalanced training data, which is another inadequacy for artificial DNNs. Numerous CT and X-ray pictures, for instance, may be utilized to evaluate negative training samples prior to the 2019 discovery of coronavirus. Conversely, the number of positive samples is rather low. Such erroneous information has the potential to impede the performance of existing deep artificial neural networks. So, the purpose of this research is to give a complete evaluation of cutting-edge deep learning frameworks for COVID-19 classifications using datasets that are publicly accessible in order to be of assistance to CAD systems.
\par
The remaining sections are structured as follows: Section II of the study reviews pertinent research. The proposed methodology is described in Section III. Section IV details the experiments and results. Sections V and VI deal with the discussion and conclusion, respectively.

\section{Related Work}
Numerous AI-based methods have been used for coronavirus classification. Consequently, the objective of this literature review is to offer a breakdown of ML and DL implementations for coronavirus complications. It will also go over some of the previously discussed topics and techniques. The experiments were done with Computed tomography and X-ray images that were publicly available.

\par

Research on 16 pre-trained Convolutional Neural Networks was proposed \cite{pham2020comprehensive} on the wide-ranging public collection of CT images from COVID-19 and healthy victims. The outcomes were presented with a mean accuracy of 93\% \& 96\% with and without data augmentation respectively. In \cite{ozturk2020automated}, an exclusive framework was proposed called the Darknet model by T.Ozturk providing diagnostics for binary \& multi-class classification using (YOLO) as a classifier. The model produced 98\% \& 87\% accuracy  for binary \& multi-class cases respectively.
\par
Asif Iqbal Khan gathered X-ray images of COVID-19 and other chest infections to generate a dataset from two publicly available datasets and propose a model called Coronet using Xception architecture \cite{khan2020coronet}. The recommended model has an overall accuracy of 89.0\%, 95.0\%, and 99.0\% for 4-class, 3-class, and binary instances, respectively. A study was carried out \cite{mahmud2020covxnet} based on DCNN titled as CovXNet with the help of efficient depth-wise convolution. Extensive experimentation with various datasets offers very adequate detecting performance with an accuracy of 97.4\% \& 90.2\% for binary \& multi-class respectively. 
\par
Using pre-trained models of Tensor flow \& Keras library, a Multi-Source Transfer Learning \cite{martinez2020classification} was proposed building on conventional Transfer Learning for classification. The models outperformed baseline ImageNet fine-tuned models thanks to their multi-source fine-tuning strategy. Moreover, the researcher also suggested an unsupervised label creation method for their Deep Residual Networks, which improves their efficiency Reaching an 89\% accuracy rate. In \cite{liu2020fast}, an automated diagnosis of coronavirus disease was developed using a lesion-attention deep neural network (LA-DNN). Two forms of valuable information were extracted: One is an indicator for whether a COVID-19 case is positive or negative, and the other is a description of positive-case of CT imaging lesions. According to the findings of the experiments, the area under the curve, sensitivity, specificity, and accuracy was 94\%, 88.8\%, 87.9\%, and 88\% respectively which meet the clinical criteria for realistic application. 
\par
A DL system termed CG-Net was suggested and assessed using 5,856 pneumonia chest X-ray pictures from a publicly accessible dataset \cite{yu2021cgnet}. Their model achieved 0.98\% accuracy, 1\% sensitivity, and 0.97\% specificity. For automated diagnosis of coronavirus illness, a simple DL model (SPPCOVID-Net) with 94\% accuracy and the lowest standard deviation across training folds. \cite{abdani2020lightweight}.
\par
Prabira Kumar \cite{sethy2020detection} developed a deep feature and SVM-based technique for automated diagnosis of coronavirus with the help of X-ray images by feeding 13 pre-trained CNN Models into the SVM classifier one by one using feature extraction. The proposed classification model has a 95.33\% accuracy rate. Research presented in \cite{wang2020covid}, COVID-Net is an open-source DCNN designed to identify COVID-19 instances from chest X-ray (CXR) pictures. The model was trained using COVIDx which is an open-access benchmark dataset comprising 13,975 CXR images from 13,870 patient cases and the most COVID-19-positive cases. In contrast to VGG-19 and Resnet-50, the model's 94\% accuracy with VGG-19 and Resnet-50 was the greatest.

\par
While most deep learning architectures have outperformed traditional methods on the COVID-19 image classification dataset, there is a lack of in-depth studies that directly compare these algorithms to the competition.
Furthermore, researchers have been working on a variety of private datasets since the inception of COVID-19; nonetheless, there is a significant possibility for research collaboration and in-depth analysis. As a consequence of this, the purpose of this research is to provide an exhaustive comparison of cutting-edge deep learning architectures for COVID-19 classifications in order to make CAD systems more accessible.

\section{Proposed Methodology}

The architectures utilized for classification, as well as details about the datasets utilized and data augmentation, are presented in this section.

\subsection{Datasets}
Four major COVID-19 X-ray and CT scan databases were used for image classification. Fig. 1 exhibits representative CT and X-ray images from these datasets.

\begin{itemize}
    \item \textbf{SARS-CoV-2:} Eduardo Soares et al. have released the SARS-CoV-2 dataset, which has a total of 2481 CT scans. Data has been gathered from COVID-19 patients in Sao Paulo, Brazil, hospitals \cite{sarscovii}.

    \item \textbf{COVID CT:} Xingyi Yang et al. have collected CT images that include 349 CT scans of 216 individuals, who tested positive for COVID-19 and 195 non-COVID-19 CT scans \cite{covidctds}.

    \item \textbf{IEEE-8023:} Joseph Paul Cohen et al have published the first publicly available COVID-19 X-ray image database \cite{ieee8023}. COVID-19 images and prognosis data from hundreds of frontal X-rays make this dataset a valuable resource for developing and accessing COVID-19 diagnosis technologies. There are 435 COVID and 505 non-COVID photos from 412 individuals in 26 countries.

    \item \textbf{Radiography Database:} Researchers from Dhaka University, Qatar University, Malaysia, and Pakistan collaborated to build a dataset of X-ray pictures of the lungs that included COVID-19 positive cases, as well as normal and viral pneumonia images \cite{radds}. As fresh X-rays of people infected with COVID-19 become available, the pictures in the database are updated appropriately.

\end{itemize}




\begin{figure}[htbp]
\centerline{\includegraphics{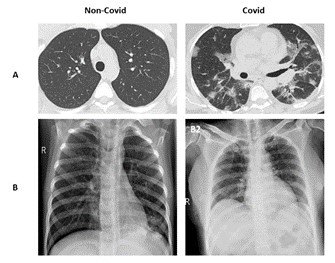}}
\caption{COVID vs Non-COVID CT \& X-ray Image}
\label{fig}
\end{figure}

\subsection{Data Augmentation}
On the training dataset, we used a variety of data augmentation strategies such as rotating the data, flipping, scaling, shifting, and shear ranging. That is, each picture in the dataset used for training is subjected to a random process. The information shown in Table I pertains to the total number of testing and training images used.

\begin{table}[htbp]
\centering
\caption{Dataset Train Test Split Information}
\begin{tabular}{|c|c|c|c|} 
\hline
\textbf{Dataset}                                                       & \textbf{Number of Images} & \textbf{Training~} & \textbf{Testing~}  \\ 
\hline
\begin{tabular}[c]{@{}c@{}}SARS COV II\\(CT) \cite{sarscovii}\end{tabular}              & 2481                      & 1860               & 621                \\ 
\hline
\begin{tabular}[c]{@{}c@{}}COVID DS  \\(CT) \cite{covidctds} \end{tabular}                & 544                       & 408                & 136                \\ 
\hline
\begin{tabular}[c]{@{}c@{}}IEEE 8023  \\(X-ray) \cite{ieee8023}\end{tabular}            & 940                       & 752                & 188                \\ 
\hline
\begin{tabular}[c]{@{}c@{}}Radiography Database \\(X-ray) \cite{radds}\end{tabular} & 15153                     & 13636              & 1517               \\
\hline
\end{tabular}
\end{table}

\subsection{Proposed Architecture}
There are numerous deep learning algorithms, the most well-known of which are CNN and RNN, researchers favor CNN for image classification. This is because CNNs are meant to take an image as input and output a variable value, which might be a probability value or a class label. As a result, researchers have employed CNN architectures to address the image classification problem. The CNN architectures used in our work as backbones for COVID-19 classification, with variants shown in parenthesis are \textit{VGG (16, 19), Densenet (121), Resnet (50, 50 V2, 101 V2),
 Mobile net (V2), Xception Inception (V3, Resnet V2),
 Efficient net (B0), and Nasnet (Large).}
\par

\par
VGG \cite{simonyan2014very} is named after the Visual Geometry Group and came in second place in the 2014’s ImageNet Competition \cite{deng2009imagenet}. It was one of the first models to demonstrate the relevance of depth in deep learning, and its simple repeating structure makes it ideal for tasks such as feature extraction. Rennets \cite{he2015deep} has enabled the training of deep networks employing layers that learn residual functions concerning layer inputs, while Densenet \cite{huang2017densely} employs feed-forward connections between each layer and every other layer.

\par

Xception \cite{chollet2016xception} is an Inception Network variant in which the inception modules have been replaced by depth-wise convolutions followed by a point-wise convolution, while inception networks \cite{szegedy2015going} are utilized to enable more efficient computation and deeper net-works by reducing dimensionality. The modules are intended to address concerns such as computational cost and overfitting. Efficient Net \cite{tan2019efficientnet} is an enhancement to Mobile Net that proposes the compound scaling module as an effective approach to evenly scale depth, breadth, and resolution. 

\par

Alternatively, Mobile Net \cite{howard2017mobilenets} was designed to meet the requirement for training and inference on devices. Finally, Nasnet \cite{nasnetl} is an algorithm that is considered to do the best on a certain task. Nasnet has been used to create networks that perform as well as or better than hand-designed systems. Methods for NAS may be classified based on the search space, search method, and performance estimation strategy employed. 

\par

The specifics of the deep learning network hyperparameters that were utilized for COVID-19 classification are detailed in Table II, and Figure 2 illustrates the architecture that is proposed for the COVID-19 image classification.


\begin{figure}
  \includegraphics[width=\textwidth,width=9.4cm]{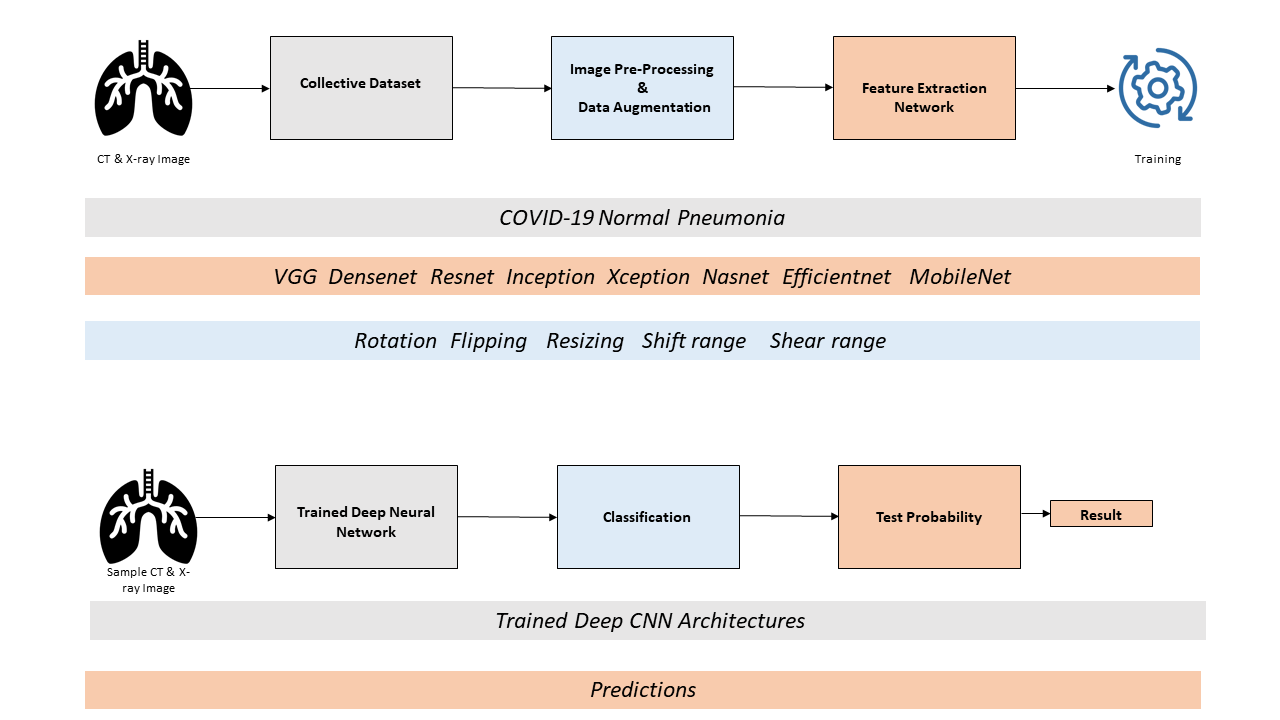}
  \caption{Proposed Architecture for Image Classification}
\end{figure}

\begin{table}[htbp]
\centering
\caption{Model Hyperparameters of Deep Learning Architectures}
\begin{tabular}{|l|l|l|l|} 
\hline
\textbf{Model }     & \textbf{Size} & \textbf{Parameters } & \textbf{Depth}  \\ 
\hline
MobileNet-V2        & 14            & 3,538,984            & 88              \\ 
\hline
EfficientNet-B0     & 29            & 5,330,571            & -               \\ 
\hline
DenseNet-121        & 33            & 8,062,504            & 121             \\ 
\hline
Xception            & 88            & 22,910,480           & 126             \\ 
\hline
Inception-V3        & 92            & 23,851,784           & 159             \\ 
\hline
ResNet 50-V2        & 98            & 25,613,800           & -               \\ 
\hline
ResNet-50           & 98            & 25,636,712           & -               \\ 
\hline
ResNet 101-V2       & 171           & 44,675,560           & -               \\ 
\hline
Inception ResNet-V2 & 215           & 55,873,736           & 572             \\ 
\hline
Nasnet Large        & 343           & 88,949,818           & -               \\ 
\hline
VGG-16              & 528           & 138,357,544          & 23              \\ 
\hline
VGG-19              & 549           & 143,667,240          & 26              \\
\hline
\end{tabular}
\end{table}

\section{Experiments \& Results}

\subsection{Model Hyperparameters}
Deep learning architectures are implemented using the Tensor flow and Keras packages. Keras supplies ImageNet-based weights for use with these pre-trained models. While there is no guarantee that the images in the ImageNet dataset, on which these models are trained, will be similar to the images gathered for research, transfer learning may be utilized to increase the effectiveness of the process.
The images were also augmented with data augmentation, and epochs were set to 20. At a learning rate of 0.0001, the batch size was set to 32 and the loss function was optimized using an ADAM optimizer. The Sigmoid activation function was employed for binary classification, whereas the SoftMax activation function was used for multi-class classification. For binary classification, the image's resolution was 224 x 224, while for multi-class classification, it was 240 x 240.
The Python scripts were run on Collaboratory, a Google-built online Jupyter Notebook environment, and Tesla K80 GPU from Google Cloud was utilized to enhance processing time.

\subsection{Evaluation Metrics}
Accuracy, precision, recall, and the F1-score were used in the appraisal of the architecture's performance.
\par
\textbf{Recall} is a measure of how well the model or algorithm predicts True Positives:
\begin{equation}
Recall = \frac{TP}{TP+FN}
\label{eq:recall}
\end{equation}

\textbf{Precision} is determined by the ratio of properly identified true negative samples to the total number of outcomes, which includes both true negative and false positive results.
\begin{equation}
Precision = \frac{TN}{TN+FP}
\label{eq:precision}
\end{equation}

\textbf{Accuracy} of a model is determined by the proportion of its predictions that are confirmed by testing.
\begin{equation}
Accuracy = \frac{TP+TN}{TP+TN+FP+FN}
\label{eq:accuracy}
\end{equation}

\textbf{F1-Score} is a technique that is used to combine the accuracy and recall of the model, and it is also the harmonic combination of the model's recall and precision.
\begin{equation}
F1 Score = \frac{2TP}{2TP+FP+FN}
\label{eq:f1}
\end{equation}

\subsection{Results}

For binary class classification, we utilized 9 deep learning architectures and implemented them on three COVID-19 binary class datasets. A closer inspection of Table IV reveals that DL architectures are reliable for diagnosing COVID-19. Resnet 101 V2 emerges the best on the “COVID CT and IEEE-8023” dataset with an accuracy of 96\% and 97\% respectively. VGG-16 architecture emerges the best on the “SARS COV II” dataset with maximal accuracy of 98\%. Similarly, on the COVID-19 multi-class “Radiography” dataset, we utilized 12 deep learning architectures and compared them with various state-of-the-art approaches. Again, the Resnet architecture with model "Resnet 50 V2" performs best with a maximum accuracy of 95\%.
The results reveal that in binary and multi-class classification, the Resnet framework significantly outperforms other deep learning architectures in terms of F1 score, accuracy, recall, and precision. Table III illustrates the confusion matrices describing the performance of classification models (Resnet-50, Resnet 101-V2, and VGG-16) on the test dataset and Table IV illustrates the major findings of the conducted experiments. 

\begin{table}
\centering
\caption{Confusion Matrices for COVID-19 Classification on different Datasets}
\begin{tabular}{|c|c|c|c|c|c|} 
\hline
\multicolumn{2}{|c|}{\begin{tabular}[c]{@{}c@{}}\textbf{SARS COV II} \\\textit{(COVID vs Normal)}\end{tabular}} & \multicolumn{2}{c|}{\begin{tabular}[c]{@{}c@{}}\textbf{IEEE-8023} \\\textit{(COVID vs Normal)}\textbf{}\end{tabular}} & \multicolumn{2}{c|}{\begin{tabular}[c]{@{}c@{}}\textbf{COVID CT DS}\\\textit{(COVID vs Normal)}\textbf{}\end{tabular}}  \\ 
\hline
303 & 10                                                                                                        & 87 & 0                                                                                                                & 83 & 4                                                                                                                  \\ 
\hline
3   & 305                                                                                                       & 6  & 95                                                                                                               & 2  & 47                                                                                                                 \\ 
\hline
\multicolumn{6}{|c|}{\begin{tabular}[c]{@{}c@{}}\textbf{Radiography DS}\\\textit{(COVID vs Normal vs Pneumonia)}\textbf{}\end{tabular}}                                                                                                                                                                                                                           \\ 
\hline
\multicolumn{2}{|c|}{293}                                                                                       & \multicolumn{2}{c|}{69}                                                                                               & \multicolumn{2}{c|}{0}                                                                                                  \\ 
\hline
\multicolumn{2}{|c|}{35}                                                                                        & \multicolumn{2}{c|}{984}                                                                                              & \multicolumn{2}{c|}{1}                                                                                                  \\ 
\hline
\multicolumn{2}{|c|}{1}                                                                                         & \multicolumn{2}{c|}{19}                                                                                               & \multicolumn{2}{c|}{115}                                                                                                \\
\hline
\end{tabular}
\end{table}

\begin{table*}
\centering
\caption{Evaluation Metrics for COVID-19 Classification}
\begin{tabular}{|p{2.5cm}|p{2.5cm}|p{2.5cm}|p{2.5cm}|p{2.5cm}|p{2.5cm}|}
\hline
\textbf{Datasets}                                                                                                                        & \textbf{Architecture}   & \textbf{Accuracy} & \textbf{Precision} & \textbf{Recall} & \textbf{F1-Score}  \\ 
\hline
\multirow{9}{*}{\begin{tabular}[c]{@{}c@{}}\textbf{IEEE 8023} \\\textbf{X-ray (Binary class)}\end{tabular}}                              & VGG-16                  & 0.94              & 0.94               & 0.94            & 0.94               \\ 
\cline{2-6}
                                                                                                                                         & VGG-19                  & 0.90              & 0.91               & 0.91            & 0.90               \\ 
\cline{2-6}
                                                                                                                                         & Xception                & 0.91              & 0.92               & 0.91            & 0.91               \\ 
\cline{2-6}
                                                                                                                                         & Resnet 50 V2            & 0.93              & 0.92               & 0.93            & 0.93               \\ 
\cline{2-6}
                                                                                                                                         & Inception V3            & 0.90              & 0.92               & 0.90            & 0.90               \\ 
\cline{2-6}
                                                                                                                                         & \textbf{Res Net 101 V2} & \textbf{0.97}     & \textbf{0.97}      & \textbf{0.97}   & \textbf{0.97}      \\ 
\cline{2-6}
                                                                                                                                         & Dense Net 121           & 0.89              & 0.89               & 0.90            & 0.89               \\ 
\cline{2-6}
                                                                                                                                         & Resnet 50               & 0.60              & 0.79               & 0.56            & 0.48               \\ 
\cline{2-6}
                                                                                                                                         & Efficient net B0        & 0.60              & 0.79               & 0.58            & 0.49               \\ 
\hline
\multirow{9}{*}{\begin{tabular}[c]{@{}c@{}}\textbf{SARS COV II}\\\textbf{~CT (Binary class)}\end{tabular}}                               & Xception                & 0.93              & 0.88               & 0.99            & 0.93               \\ 
\cline{2-6}
                                                                                                                                         & Inception V3            & 0.96              & 0.99               & 0.93            & 0.96               \\ 
\cline{2-6}
                                                                                                                                         & Dense Net 121           & 0.95              & 0.98               & 0.93            & 0.96               \\ 
\cline{2-6}
                                                                                                                                         & \textbf{VGG-16}         & \textbf{0.98}     & \textbf{0.97}      & \textbf{0.99}   & \textbf{0.98}      \\ 
\cline{2-6}
                                                                                                                                         & VGG-19                  & 0.96              & 0.98               & 0.94            & 0.96               \\ 
\cline{2-6}
                                                                                                                                         & ResNet-50               & 0.80              & 0.81               & 0.80            & 0.80               \\ 
\cline{2-6}
                                                                                                                                         & Resnet50v2              & 0.96              & 0.95               & 0.97            & 0.96               \\ 
\cline{2-6}
                                                                                                                                         & Resnet101v2             & 0.95              & 0.97               & 0.93            & 0.95               \\ 
\cline{2-6}
                                                                                                                                         & Efficient net B0        & 0.50              & 1.00               & 0.50            & 0.67               \\ 
\hline
\multirow{9}{*}{\begin{tabular}[c]{@{}c@{}}\textbf{COVID DS}\\\textbf{CT (Binary class)}\end{tabular}}                                   & VGG-16                  & 0.90              & 1.00               & 0.87            & 0.93               \\ 
\cline{2-6}
                                                                                                                                         & VGG-19                  & 0.90              & 0.98               & 0.88            & 0.92               \\ 
\cline{2-6}
                                                                                                                                         & Xception                & 0.92              & 0.85               & 0.96            & 0.90               \\ 
\cline{2-6}
                                                                                                                                         & Resnet 50 V2            & 0.92              & 0.98               & 0.90            & 0.94               \\ 
\cline{2-6}
                                                                                                                                         & Inception V3            & 0.89              & 0.95               & 0.88            & 0.92               \\ 
\cline{2-6}
                                                                                                                                         & \textbf{Res Net 101 V2} & \textbf{0.96}     & \textbf{0.95}      & \textbf{0.98}   & \textbf{0.97}      \\ 
\cline{2-6}
                                                                                                                                         & Dense Net 121           & 0.93              & 0.97               & 0.92            & 0.94               \\ 
\cline{2-6}
                                                                                                                                         & Efficient Net B0        & 0.64              & 1.00               & 0.64            & 0.78               \\ 
\cline{2-6}
                                                                                                                                         & Resnet 50               & 0.85              & 0.97               & 0.83            & 0.89               \\ 
\hline
\multirow{12}{*}{\begin{tabular}[c]{@{}c@{}}\textbf{COVID-19 }\\\textbf{Radiography DS }\\\textbf{X-ray (Multi Class)}\end{tabular}} & Xception                & 0.87              & 0.88               & 0.82            & 0.84               \\ 
\cline{2-6}
                                                                                                                                         & Inception V3            & 0.85              & 0.89               & 0.70            & 0.75               \\ 
\cline{2-6}
                                                                                                                                         & Nasnet Large            & 0.89              & 0.87               & 0.87            & 0.87               \\ 
\cline{2-6}
                                                                                                                                         & \textbf{ResNet-50 V2}   & \textbf{0.95}     & \textbf{0.93}      & \textbf{0.88}   & \textbf{0.90}      \\ 
\cline{2-6}
                                                                                                                                         & Dense Net 121           & 0.89              & 0.91               & 0.84            & 0.87               \\ 
\cline{2-6}
                                                                                                                                         & Mobile Net V2           & 0.91              & 0.90               & 0.89            & 0.89               \\ 
\cline{2-6}
                                                                                                                                         & VGG-16                  & 0.80              & 0.83               & 0.67            & 0.69               \\ 
\cline{2-6}
                                                                                                                                         & VGG-19                  & 0.79              & 0.80               & 0.69            & 0.72               \\ 
\cline{2-6}
                                                                                                                                         & Inception Resnet V2     & 0.90              & 0.91               & 0.80            & 0.84               \\ 
\cline{2-6}
                                                                                                                                         & Efficient net B0        & 0.67              & 0.70               & 0.68            & 0.65               \\ 
\cline{2-6}
                                                                                                                                         & Resnet 50               & 0.81              & 0.78               & 0.89            & 0.81               \\ 
\cline{2-6}
                                                                                                                                         & Resnet 101 V2           & 0.91              & 0.94               & 0.85            & 0.89               \\
\hline
\end{tabular}
\end{table*}



\section{Discussion}
The lack of comprehensive experiments and comparisons with basic baselines in the enormous number of novel models offered is one rationale for our work. This study may be used as a collection of architectures that DL model designers can use to check and assess if their innovative model outperforms others, such as by comparing their metrics to models with the same number of parameters and/or training/validation dataset. 
We used a variety of alternative deep learning architectures for the COVID-19 imaging classification task, which allowed us to classify both X-ray and CT pictures. When compared to the outcomes of other deep learning architectures, the Resnet framework shows substantial improvements in terms of accuracy, precision, recall, and F1 score. Table V provides a comparison of the suggested models with several cutting-edge classification methods that have been discussed in the research literature.

\begin{table}[htbp]
\centering
\caption{Comparison of the proposed methodology to several cutting-edge baseline classification approaches}
\begin{tabular}{|c|c|c|} 
\hline
\textbf{Dataset}                                                                                 & \textbf{Author}   & \begin{tabular}[c]{@{}c@{}}\textbf{Result}\\\textbf{(Accuracy)}\end{tabular}  \\ 
\hline
\multirow{4}{*}{\textbf{SARS COV II }}                                                           & Nguyen [26]       & 0.83                                                                          \\ 
\cline{2-3}
                                                                                                 & Hasan [27]        & 0.78                                                                          \\ 
\cline{2-3}
                                                                                                 & Martinez [7]      & 0.87                                                                          \\ 
\cline{2-3}
                                                                                                 & \textbf{Proposed} & \textbf{0.98}                                                                 \\ 
\hline
\multirow{4}{*}{\textbf{COVID CT}}                                                               & Panwar [28]       & 0.94                                                                          \\ 
\cline{2-3}
                                                                                                 & Wang~ [29]        & 0.90                                                                          \\ 
\cline{2-3}
                                                                                                 & Amyar [30]        & 0.94                                                                          \\ 
\cline{2-3}
                                                                                                 & \textbf{Proposed} & \textbf{0.96}                                                                 \\ 
\hline
\multirow{4}{*}{\textbf{IEEE-8023}}                                                              & Horry [31]        & 0.79                                                                          \\ 
\cline{2-3}
                                                                                                 & Arellano [32]     & 0.94                                                                          \\ 
\cline{2-3}
                                                                                                 & Militante [33]    & 0.95                                                                          \\ 
\cline{2-3}
                                                                                                 & \textbf{Proposed} & \textbf{0.97}                                                                 \\ 
\hline
\multirow{3}{*}{\begin{tabular}[c]{@{}c@{}}\textbf{Radiography}\\\textbf{Database}\end{tabular}} & Abdani\cite{abdani2020lightweight}            & 0.94                                                                          \\ 
\cline{2-3}
                                                                                                 & Fangoh\cite{fangoh2020using}            & 0.82                                                                          \\ 
\cline{2-3}
                                                                                                 & \textbf{Proposed} & \textbf{0.95}                                                                 \\
\hline
\end{tabular}
\end{table}

\section{Conclusion}
In order to simplify CAD systems, we have offered a detailed comparison of current cutting-edge DL classification architectures for COVID-19. The first is a binary classification that seeks to distinguish between CT and X-ray pictures of COVID-19 and non-COVID-19 instances. Then, we classified COVID-19, normal, and viral pneumonia X-ray images using a multi-class classification scheme. It reaches a maximum accuracy of 98\% for binary class \& 95\% for multi-class classification. In the future, the effect of higher input resolution can be explored and we can use the proposed framework with other COVID-19 datasets as it becomes available. To conclude, the results demonstrate that deep learning architectures can help in CAD systems and can assist radiologists in correctly and fastly detecting COVID-19 using radiological imaging in less time.

\end{document}